\pdfoutput=1
\PassOptionsToPackage{table,usenames,dvipsnames,svgnames}{xcolor}
\documentclass[11pt]{article}

\usepackage[final]{acl}
\usepackage{amssymb}
\usepackage{amsthm}
\usepackage{enumitem}
\usepackage{color}
\usepackage{url}
\usepackage{tabularx}

\usepackage{times}
\usepackage{latexsym}
\usepackage{makecell}
\usepackage{booktabs}
\usepackage{CJKutf8}
\usepackage{multirow}
\usepackage{graphicx}
\usepackage{amsmath}
\usepackage[T1]{fontenc}
\usepackage{pifont}
\usepackage{setspace}
\usepackage[most]{tcolorbox}
\usepackage[utf8]{inputenc}
\usepackage[linesnumbered,lined,ruled,commentsnumbered]{algorithm2e}
\usepackage{bm}
\usepackage{microtype}

\usepackage{inconsolata}
\usepackage{cleveref}
\usepackage{cancel}

\crefname{section}{§}{§§}
\Crefname{section}{§}{§§}

\definecolor{amaranth}{rgb}{0.9, 0.17, 0.31}
\definecolor{kellygreen}{rgb}{77, 186, 23}
\definecolor{azure}{rgb}{0.0, 0.5, 1.0}
\definecolor{gred}{rgb}{0.9, 0.17, 0.31}
\definecolor{gblue}{rgb}{0.0, 0.5, 1.0}
\definecolor{gyellow}{RGB}{244,180,0}
\definecolor{ggreen}{rgb}{0.3, 0.73, 0.09}
\definecolor{ggrey}{RGB}{115,115,115}
\definecolor{Emerald}{rgb}{0.004, 0.804, 0.004}

\newcommand{\cmark}{{\color[HTML]{00715E}{\ding{51}}}}
\newcommand{\xmark}{{\color[HTML]{961C26}{\ding{55}}}}

\newcommand{\start}[1]{\vspace{.6mm}\noindent{{\bf #1}\ }}

\title{Incomplete Utterance Rewriting with Editing Operation Guidance and Utterance Augmentation}

\author{Zhiyu Cao, Peifeng Li\thanks{ \ \ Corresponding author}, Yaxin Fan, Qiaoming Zhu \\
        School of Computer Science and Technology, Soochow University, Suzhou, China  \\
        \texttt{zycao18@stu.suda.edu.cn, yxfansuda@stu.suda.edu.cn} \\
        \texttt{\{pfli, qmzhu\}@suda.edu.cn}
        }

\begin{document}
\maketitle
\begin{abstract}
Although existing fashionable generation methods on Incomplete Utterance Rewriting (IUR) can generate coherent utterances, they often result in the inclusion of irrelevant and redundant tokens in rewritten utterances due to their inability to focus on critical tokens in dialogue context. Furthermore, the limited size of the training datasets also contributes to the insufficient training of the IUR model. To address the first issue, we propose a multi-task learning framework EO-IUR (Editing Operation-guided Incomplete Utterance Rewriting) that introduces the editing operation labels generated by sequence labeling module to guide generation model to focus on critical tokens. Furthermore, we introduce a token-level heterogeneous graph to represent dialogues. To address the second issue, we propose a two-dimensional utterance augmentation strategy, namely editing operation-based incomplete utterance augmentation and LLM-based historical utterance augmentation. The experimental results on three datasets demonstrate that our EO-IUR outperforms previous state-of-the-art (SOTA) baselines in both open-domain and task-oriented dialogue. The code will be available at \url{https://github.com/Dewset/EO-IUR}.
\end{abstract}

\section{Introduction}
To express concisely and conveniently in multi-turn dialogues, speakers tend to use incomplete utterances which usually omit (i.e., ellipsis) or refer back (i.e., coreference) to the history of dialogue context. 
\citet{su-etal-2019-improving} reported that more than 70\% of utterances exhibited the phenomena of coreference and ellipsis, particularly in pro-drop languages like Chinese.
This presents a significant challenge for the application of natural language understanding, particularly in the context of virtual assistants and customer support systems \cite{hauswald2015sirius, debnath2018identifying}. 
To address this issue, the task of Incomplete Utterance Rewriting (IUR) is proposed to generate complete utterances that can
be understood by AI systems without any additional context.
In particular, IUR is used to perform ellipsis and coreference resolution in dialogues. Taking Table ~\ref{rewrite-examples} as an example, ``He'' in the utterance $u_3$ refers to ``Tolstoy'' in the historical utterances and ``Anna Karenina'' is omitted in $u_3$. Hence, the rewritten utterance of $u_3$ would be $u_3^{\prime}$.

\begin{table}[t]
\centering
\resizebox{\linewidth}{!}{
\begin{tabular}{cc}
\toprule
\textbf{Speaker (turn)} & \textbf{Utterance}\\
\midrule
$\mathbf{Speaker_1(u_1)}$ & \makecell[c]{Do you know \textcolor[RGB]{248,122,23}{Anna Karenina}?}  \\
\midrule
$\mathbf{Speaker_2(u_2)}$ & \makecell[c]{Who is \textcolor{blue}{Tolstoy}?} \\
\midrule
$\mathbf{Speaker_1(u_3)}$ & \makecell[c]{He is the author.}  \\
\midrule
$\mathbf{Speaker_1(u_3^{\prime})}$ & \makecell[c]{\textcolor{red}{\bcancel{He}} \textcolor{blue}{\textit{Tolstoy}} is the author of \textcolor[RGB]{248,122,23}{Anna Karenina}.}  \\
\bottomrule
\end{tabular}
}
\caption{\label{rewrite-examples}
An example of IUR. The first two utterances are historical utterances, $u_3$ is an incomplete utterance, and $u_3^{\prime}$ is the rewritten utterance.
}
\vspace{-0.6cm}
\end{table}

Most current IUR models use generation \citep{su-etal-2019-improving, zhou-etal-2019-unsupervised, xu-etal-2020-semantic} or sequence labeling paradigms \citep{DBLP:conf/aaai/JinSJ0G22, si-etal-2022-mining, chen-2023-incomplete, du-etal-2023-multi, li-etal-2023-well}. Each paradigm has its own set of advantages and disadvantages. 
Sequence labeling methods commonly suffer from two main issues: grammatical errors and incomplete rewritten utterances. These issues arise due to missing relative insertion positions when multiple tokens are inserted into one position and imbalanced positive and negative samples (i.e., tokens that do not need to be modified or inserted), respectively.
The generation methods suffer from generating redundant tokens that are irrelevant to the dialogue context due to the lack of focus on critical tokens in dialogue context (e.g., ``Anna Karenina'' and ``Tolstoy'' in $u_1$ and $u_2$ ).
Only a few  methods  \cite{DBLP:conf/aaai/HuangLZ021,inoue-etal-2022-enhance} attempt to combine the above two methods to overcome their respective shortcomings. However, they merely utilise a multi-task learning framework to combine sequence labeling and generation methods. This approach lacks direct interaction between the two methods, in which sequence labeling is unable to effectively guide generation to focus on the critical tokens in dialogue context in order to avoid generating redundant tokens.

Furthermore, the majority of annotated IUR datasets are relatively small in size (e.g., the size of CQR \cite{regan2019dataset} is only 0.64K). Despite the existence of numerous data augmentation methods in NLP, they are designed for ordinary documents. The structure of dialogue is more complex and personal pronouns appear with greater frequency. However, there is no research on data augmentation in IUR.

To address the first issue, we propose a multi-task learning framework, EO-IUR (Editing Operation-guided Incomplete Utterance Rewriting), that incorporates editing operation labels generated by the sequence labeling model to guide the generation model. This allows the decoder of the generation model to focus on the critical tokens in historical utterances and incomplete utterances, with the use of four types of defined editing operations. Furthermore, we introduce a token-level heterogeneous graph to represent dialogues, which enables the model to learn the syntactic structure corresponding to the omitted tokens and the relationships between coreference-related tokens.

To address the second issue, we propose a two-dimensional utterance augmentation strategy for IUR, namely editing operation-based incomplete utterance augmentation and LLM-based historical utterance augmentation. These are used to augment incomplete utterances and the historical utterances, respectively.
The experimental results on two Chinese datasets and one English dataset show that our proposed model outperforms several SOTA baselines significantly.

\section{Related Work}

\start{Sequence Labeling Methods}
  A rewritten utterance often maintains a high degree of structural and content similarity with the corresponding incomplete utterance to be rewritten. Therefore, most of the generated content comes from the context of the dialogue, leading to the emergence of numerous sequence labeling-based methods. To address the issue of low coverage during IUR using sequence labeling, \citet{DBLP:conf/aaai/JinSJ0G22} proposed a hierarchical context marker to address this issue. \citet{si-etal-2022-mining} explicitly introduced semantic structured information through carefully designed inquiry templates. To better extract information from the dialogue context, \citet{chen-2023-incomplete} identified spans in the context and their order, and then combined them into rewritten utterance. \citet{du-etal-2023-multi} proposed a multi-granularity information capturing framework for incomplete utterance rewriting. \citet{li-etal-2023-well} used a single-layer MLP architecture to mine latent semantic information for IUR tasks.
 
\start{Generation Methods}
The emergence of pre-trained generation models has prompted previous studies to investigate the potential of generation models in IUR. \citet{su-etal-2019-improving} enhanced the Transformer framework \citep{DBLP:conf/nips/VaswaniSPUJGKP17} by incorporating pointer networks, which allow for generating utterances by copying tokens from either the dialogue history or the current utterance. \citet{zhou-etal-2019-unsupervised} introduced a pretraining approach using pseudo-parallel data and subsequently employed reinforcement learning to optimize the reward for generating final answers. \citet{xu-etal-2020-semantic} introduced semantic role labeling into the generation process of IUR to provide additional information. 

\start{Combination of Generation and Sequence Labeling Methods} 
Only a few studies have focused on the combination of generation and sequence labeling methods. \citet{DBLP:conf/aaai/HuangLZ021} set the state of each token in the incomplete utterance to be reserved, replaced, or deleted. However, they did not consider the state of tokens in the dialogue history. \citet{inoue-etal-2022-enhance} only focused on missing content in the current utterance without considering its position. Furthermore, although they performed joint training for sequence labeling and generation, they did not consider the intrinsic connection between these two paradigms and treated them as independent. 
Our proposed method differs from the aforementioned approachs in that it incorporates a multi-task learning framework, which enables the direct interaction between the sequence labeling and generation methods. This framework incorporates editing operation labels generated by sequence labeling module, guiding the generation model and allowing the decoder to focus on the critical tokens in dialogue context. 

\section{Methodology}

\subsection{Task Formulation and Overview}

\begin{figure*}[ht]
\begin{center}
 \includegraphics[width=1\linewidth]{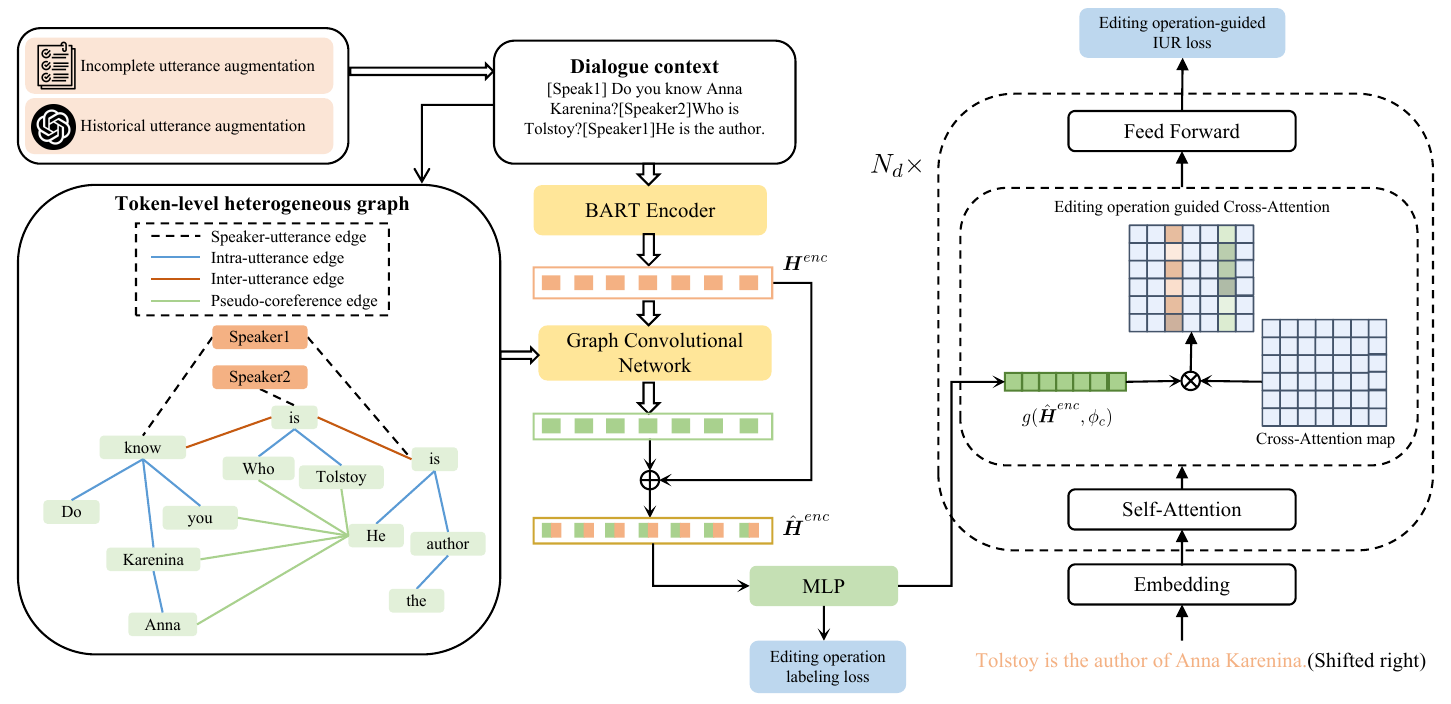}
 \caption{Overview of our model EO-IUR, which includes  utterance augmentation, construction of token-level heterogeneous graph convolutional neural network, editing operation labeling, and editing operation-guided IUR.}
 \label{fig:method}
\end{center}
\end{figure*}

Given a dialogue  $\mathcal{D}=\left\{s_i\right\}^{N}_{i=1}$ containing $N$ utterances where $s_i=\left\{w_j\right\} ^{M}_{j=1}$ refers to the $i$-th utterance containing $M$ tokens, $s_N$ is the incomplete utterance, and $\mathcal{T}=\left\{w_j\right\}^{L}_{j=1}$ containing $L$ tokens is the rewritten utterance. Providing $\mathcal{D}$ as input, the task of IUR is to generate the rewritten utterance $\mathcal{T}$ by modeling the conditional probability distribution $P(\mathcal{T}|\mathcal{D})$.

We view IUR here as an end-to-end editing operation-guided generation task and its overview is shown in Fig. \ref{fig:method}. Given a dialogue, we first encode the dialogue context using the encoder of a generation model (\cref{sec:fea_init}), and then we construct a token-level heterogeneous dialogue graph (\cref{sec:con_graph}) to merge the four types of information: syntax, utterance, speaker, and coreference. Subsequently, we propose a multi-task learning framework, in which we introduce a sequence labeling module to label editing operations of the tokens in the dialogue (\cref{sec:cls_gen}). Finally, we use the labels of editing operations to guide rewritten utterance generation. Furthermore, in order to address the issue of limited data size and short utterances, we propose two strategies for augmenting the existing utterances: editing operation-based utterance augmentation and LLM-based historical utterance augmentation. The former is designed for incomplete utterances, while the latter is intended for historical utterances (\cref{sec:da}).

\subsection{Feature Initialization}
\label{sec:fea_init}

The encoder of BART$_{\operatorname{base}}$ \citep{lewis-etal-2020-bart} is employed to extract features from dialogue utterances, which are used to  initialize node embeddings in our subsequent graph convolutional network (GCN). Specifically, for the $i$-th utterance, we add a special marker ``[$Speaker_i$]'' in front of it to represent the speaker, resulting in an input format of ``\{[$Speaker_1$] $s_1$ [$Speaker_2$] $s_2$,...\}'' to represent the dialogue $\mathcal{D}$. Subsequently,  the output of the final layer of the encoder is extracted as the feature representation $\boldsymbol{H}^{enc} \in \mathbb{R}^{K*d_u}$ of dialogue $\mathcal{D}$, where $K$ denotes the number of input tokens and $d_u$ represents the dimension of feature representations.

\subsection{Dialogue as a Heterogeneous Graph}
\label{sec:con_graph}
IUR involves the resolution of coreference and ellipsis, in which syntactic structures (e.g., the omission of subjects and predicates) serve as critical cues for the comprehension of the relations among tokens.  However, most of the previous work \citep{hao-etal-2021-rast, chen-2023-incomplete} on IUR directly used BERT \citep{devlin2019bert} as the encoder to extract dialogue features. BERT is primarily pre-trained on textual corpora and is therefore difficult to capture the intricate structural information in dialogues. To facilitate the learning of the syntactic structure corresponding to the omitted tokens and relations among coreference-related tokens in dialogues, we introduce graph neural networks to  further enhance dialogue information.

First, we use the dependency parsing tool spacy\footnote[1]{\url{https://github.com/explosion/spaCy}} to obtain a syntactic tree representation of each utterance in dialogues.
 The graph is represented as $\mathcal{G}=(\mathcal{V},\mathcal{E},\mathcal{R})$. Here the node set $\mathcal{V}=\mathcal{V}_{uttr} \cup \mathcal{V}_{spk}$, where $\mathcal{V}_{uttr}$ and $\mathcal{V}_{spk}$ represent  the token nodes and speaker nodes respectively; $\mathcal{E}$ is the set of edges connecting vertices in $\mathcal { V }$; and $\mathcal{R}$ denotes the set of relational types of edges as defined below. 
1)  \emph{Intra-utterance edge}: Given the significance of syntactic information for IUR, for each token in an utterance, we establish connections between adjacent nodes on the corresponding syntax tree.
 2) \emph{Inter-utterance edge}: To integrate the information from each utterance with others, we connect the root nodes of adjacent utterances' syntax trees with edges.
3) \emph{Speaker-utterance edge}: To integrate the speaker information, we connect each speaker's tag to the root nodes of their corresponding syntax trees.
4) \emph{Pseudo-coreference edge}: Coreference information is crucial for IUR. However, previous work for coreference resolution has been unable to accurately identify all coreference cases. Therefore, here we connect all pronouns in the incomplete utterance with pronouns and nouns in all other utterances.

Subsequently, we utilise $I$-layer GCNs to aggregate the features of neighbouring nodes.  The initial representation of each node is the output of its corresponding token in the encoder. Given the node $i$ in the layer $l$, the graph convolution operation is defined as follows, 
\begin{equation}
    \begin{aligned}
       \boldsymbol{s}_{i}^0 &= \boldsymbol{H}_i^{enc} \\
       \boldsymbol{s}_{i}^{(l + 1)} &= \sigma \left(\sum_{r\in\mathcal{R}}\sum_{v\in\mathcal{N}_r(i)} W^{(l)}_r \boldsymbol{s}_{v}^{(l)} + b^{(l)}_r\right)
    \end{aligned}
\end{equation}
where $\mathcal{R}$ is the set of different types of edges mentioned above, $W^{(l)}_r\in \mathbb{R}^{d_u\times d_u}$ and $b^{(l)}_r \in \mathbb{R}^{d_u}$ are trainable parameters, $\mathcal{N}_r(i)$ denotes the neighbors connected to the node $i$ via the $r$-th type of edge, and $\sigma$ represents the activation function.

For each node's features, we average them with the corresponding representations generated by the encoder to obtain the final representation $\hat{\boldsymbol{H}}^{enc}_i=\frac{\boldsymbol{s}_{i}^{I} \oplus \boldsymbol{H}_i^{enc}}{2}$.

\subsection{Editing Operation-guided IUR}
\label{sec:cls_gen}
As mentioned in introduction, IUR can be specifically categorized into two distinct operations: coreference  resolution and ellipsis resolution. Although existing pre-trained generation models demonstrate strong generation capabilities, they do not explicitly consider the two distinct operations of  replacement (for coreference) and insertion (for ellipsis) in IUR \citep{su-etal-2019-improving, xu-etal-2020-semantic, zhou-etal-2019-unsupervised}. Consequently, existing generation models frequently fail to attend to crucial information present in dialogue context, such as entities, and may also generate redundant tokens that are unrelated to the dialogue.

To address the aforementioned issue, we propose an editing operation-guided IUR method that focuses the model on the critical tokens in dialogue context, thereby enabling the generation of context-related utterances. In particular, we first introduce a sequence labeling module that generates the labels of edit operations, which are then used to guide the following utterance generation.

\noindent \textbf{Editing Operation Labeling} During the process of IUR, only a very small amount of tokens in dialogue context will be used, i.e., tokens involving substitution and insertion. 
To make the model focus more on these critical tokens, we propose a token-level sequence labeling task to generate the labels of editing operations. 

The most common editing operations are insertion, deletion, and replacement. In IUR, there are only two operations: insertion and replacement. To this end, we defined three labels: ``RP'', ``NW'', and ``IN'' to correspond to these two operations and ``NA'' to correspond to no operations. This is illustrated in Table ~\ref{label-scheme}.
The labels ``RP'' and ``NW'' involve  replacement operations in coreference resolution, while ``IN'' corresponds to insertion operations in ellipsis resolution. 

A sequence labeling module is introduced to generate editing operations, with an illustrative example provided in Table ~\ref{label-cons}. In this example, since the pronoun ``He'' refers to the entity ``Tolstoy'',  ``He'' is the token being replaced (corresponding to the ``RP'' label), and ``Tolstoy'' is the token after replacement (corresponding to the ``NW'' label). Moreover, the entity  ``Anna Karenina'' is the span that needs to be inserted (corresponding to the ``IN'' label). At this point, we do not know the specific insertion position.

\begin{table}[]
\centering
\resizebox{\linewidth}{!}{
\begin{tabular}{ll}
\toprule
\textbf{Label} & \textbf{Description}\\
\midrule
$\mathbf{NA}$ & Tokens that do not involve editing operations.  \\
$\mathbf{RP}$ & Tokens that needs to be replaced. \\
$\mathbf{NW}$ & Tokens that used to replace other token.  \\
$\mathbf{IN}$ & Tokens that need to be inserted.  \\
\bottomrule
\end{tabular}
}
\caption{
Four labels of editing operations and non-operation.
}
\label{label-scheme}
\end{table}

\begin{table}[]
\centering
\resizebox{\linewidth}{!}{
\begin{tabular}{ll}
\toprule
\textbf{Dialogue context} & \textbf{Label sequence}\\
\midrule
Do you know \textcolor{DarkGreen}{Anna Karenina}? & NA NA NA \textcolor{DarkGreen}{IN IN} NA  \\
Who is \textcolor{red}{Tolstoy}? & NA NA \textcolor{red}{NW} NA \\
\textcolor{blue}{He} is the author. & \textcolor{blue}{RP} NA NA NA NA  \\
\bottomrule
\end{tabular}
}
\caption{
An example of labels, where blue, red and green fonts corresponds to ``RP'' , ``NW'' and ``IN'' labels, respectively.
}
\label{label-cons}
\end{table}

After the input dialogue context $\mathcal{D}$ is encoded and enhanced with graph neural network features, we obtain the contextual representation $\hat{\boldsymbol{H}}^{enc}$. We feed this representation into a two-layer MLP with $\tanh (\cdot)$ activation function, and then pass it through a softmax layer to obtain the label probability distribution for each token as follows,
\begin{equation}
       g(\hat{\boldsymbol{H}}^{enc}, \phi_c)=\operatorname{softmax}(\mathbf{MLP}(\hat{\boldsymbol{H}}^{enc} ; \phi_c))
\end{equation}
where $\phi_c$ represents the MLP parameters corresponding to label $c$.
Finally we optimize the model using cross-entropy loss function as follows, 
\begin{equation}
       \mathcal{L}_{eol}=-\sum_{i=1}^{|\mathcal{C}|} y^i \log g(\hat{\boldsymbol{H}}^{enc}, \phi_i)
\end{equation}
where $y^i$ is the $i$-th label and $|\mathcal{C}|$ is the total number.

\noindent  \textbf{Editing Operation-guided Utterance Generation} To use the labels of editing operations to facilitate the utterance generation at the decoder side, we define three types of key tokens, i.e., tokens corresponding to the labels ``RP'', ``NW'', and ``IN''. We hope that the decoder can pay more attention to these tokens when generating rewritten utterances. 

To accomplish this, we modify the cross-attention layer between the encoder and decoder. This is done by multiplying the attention score by an influence factor, which is the sum of the probabilities that each input token belongs to the labels ``RP'', ``NW'' and ``IN'', as shown below, 
\begin{equation}
\begin{split}
    \boldsymbol{\lambda} &=1-g(\hat{\boldsymbol{H}}^{enc}, \phi_{0}) \\
    attn_{<j,i>}&=\frac{exp((\tau_d+\boldsymbol{\lambda}_i)\times attn_{<j,i>})}{\sum_{k=1}^K exp((\tau_d+\boldsymbol{\lambda}_k) \times attn_{<j,k>})}
\end{split}
\end{equation}
where $\phi_{0}$ represents the MLP parameters corresponding to the label ``NA'', $attn_{<j,i>}$ denotes the attention score of the $j$-th token of the output with respect to the $i$-th token of the input. To prevent the attention scores from degrading to zero due to some of the influence factors $\boldsymbol{\lambda}_i$ converging to zero, we add a temperature coefficient $\tau_d$ to them, which has a smoothing effect. By scaling the cross-attention scores as above, the decoder is made to focus more on the key tokens.

We optimize the generation of rewritten utterances by minimizing the negative log-likelihood loss function $\mathcal{L}_\text{gen}$ as follows,
\begin{equation}
\label{equation:gen-loss}
    \mathcal{L}_\text{gen} = -\log \sum_{i=1}^{L} P(\mathcal{T}_i|\mathcal{T}_{<i}, \mathcal{D}, \theta_{gen}),
\end{equation}
where $\theta_{gen}$ represents the parameters of the generation model, and $\mathcal{T}_{<i}$ represents the partially rewritten utterance that have been generated before the $i$-th token is generated.

\noindent  \textbf{Joint Optimization}
We jointly optimize the loss of editing operation labeling and rewritten utterance generation through the following objectives:
\begin{equation}
       \mathcal{L} = \alpha_1\mathcal{L}_\text{gen} + \alpha_2 \mathcal{L}_\text{eol}
\end{equation}
where $\alpha_1$ and $\alpha_2$ represent the weights of each module, respectively. To improve the stability of training, we first use only the utterance generation task to train the model for warm-up, and then add the editing operation labeling tasks.

\subsection{Utterance Augmentation}
\label{sec:da}

The majority of existing IUR datasets are relatively small, which makes it challenging to train an efficient model. Furthermore, colloquial dialogues frequently contain ellipsis and coreference, resulting in brief and ambiguous utterances. To address the above two issues, we propose two data augmentation strategies to expand existing IUR datasets: editing operation-based incomplete utterance augmentation and LLM-based historical utterance augmentation.

\noindent  \textbf{Editing Operation-based Incomplete Utterance Augmentation} In IUR, there are two cases involving ellipsis resolution and coreference resolution. It is hypothesized that these two cases can be transformed into each other, and a strategy is proposed to augment training samples. The incomplete utterance is compared with the rewritten utterance to identify the positions of ellipsis and coreference. In instances where ellipsis occurs, a pronoun is inserted to indicate coreference at that point. Conversely, for instances of coreference, the reference token is deleted, thereby converting it into an ellipsis. This approach enables bidirectional conversion between ellipsis and coreference. 

\noindent  \textbf{LLM-based Historical Utterance Augmentation}
We also consider enhancing the dialogue history. To this end, we employ a large language model (LLM) to rewrite historical utterances without altering the contextual semantics. Here, we utilize the method of in-context learning. Initially, several samples are generated through LLM in a zero-shot manner. Subsequently, the five samples with the highest rewriting quality are manually selected as examples for data augmentation purposes. Appendix ~\ref{appendix:augexample} and ~\ref{appendix:augprompt} respectively provide an example of data augmentation and the prompts used during the data augmentation process.

\section{Experimentation}

\subsection{Experimental Settings}
\start{Datasets}
To comprehensively evaluate the effectiveness of our EO-IUR, following previous work \citep{si-etal-2022-mining}, we conducted evaluation on three popular datasets: the Chinese open-domain dialogue datasets REWRITE \citep{su-etal-2019-improving} and RESTORATION-200K (RES200K) \citep{pan-etal-2019-improving}, and the English task-oriented dialogue dataset TASK \citep{quan-etal-2019-gecor}. Specific statistics for the three datasets are provided in Appendix~\ref{appendix:dataset} and the implementation details are listed in Appendix~\ref{appendix:implementation}.

\start{Evaluation Metrics}
Similar to previous work \citep{chen-2023-incomplete,li-etal-2023-incomplete}, we use the following metrics: 1) BLEU$_n$ \citep{papineni-etal-2002-bleu} measures the accuracy by calculating the degree of matching of n-grams in the generated utterances and the reference utterances, and here we use BLEU-1 (B1), BLEU-2 (B2) and BLEU-4 (B4). 2) ROUGE$_n$ \citep{lin-2004-rouge} measures the degree of overlap of n-grams in the generated and reference utterances, focusing on recall, and here we use ROUGE-1 (R1), ROUGE-2 (R2) and ROUGE-L (RL). 3) Restoration F-score$_n$ (F$_n$ for short) \citep{pan-etal-2019-improving} measures how much missing information is added to the incomplete utterance, and here we use F$_1$, F$_2$ and F$_3$. 4) Exact Match (EM) measures how many generated utterances are completely correct. 

\start{Baselines}
We compare our model EO-IUR with the following strong baselines: BART$_{\operatorname{base}}$ \citep{lewis-etal-2020-bart}, RAST \citep{hao-etal-2021-rast}, SARG \citep{DBLP:conf/aaai/HuangLZ021}, HCT \citep{DBLP:conf/aaai/JinSJ0G22}, QUEEN \citep{si-etal-2022-mining}, RAU \citep{DBLP:conf/icassp/ZhangLWCX22}, 
SGT \citep{chen-2023-incomplete}, MGIIF \citep{du-etal-2023-multi} and MIUR \citep{li-etal-2023-well}.

\begin{table}[h]
\centering
\small
\setlength{\tabcolsep}{2.8mm}
\begin{tabular}{lccccccccccccc}
\toprule

Model &  EM & B$_4$ & F$_1$  \\
\midrule
BART$_{\operatorname{base}}$ \citep{lewis-etal-2020-bart} &  70.1 & 83.9	& 69.5\\
RUN \citep{liu-etal-2020-incomplete} & 70.6 & 86.1 & 68.3\\
QUEEN \citep{si-etal-2022-mining} & 71.6 & 86.3 & NA\\
MIUR \citep{li-etal-2023-well} & 70.9 & 86.0 & 72.3\\
SGT \citep{chen-2023-incomplete} & 71.1 & 86.7 & 85.0\\
\midrule
EO-IUR (Ours) & \textbf{80.8} & \textbf{90.6} & \textbf{87.6}\\
\bottomrule
\end{tabular}
\caption{Result comparison on English TASK.}
\label{tab:task}
\end{table}

\begin{table*}[]
\centering
\small
 \setlength{\tabcolsep}{3.6mm}
\begin{tabular}{lcccccccccccccc}
\toprule

Model & F$_1$ & F$_2$  & F$_3$ & B$_1$ & B$_2$ & R$_1$ &R$_2$ & R$_L$ & EM   \\
\midrule
BART$_{\operatorname{base}}$ \citep{lewis-etal-2020-bart} & 81.2 & 76.0 & 79.7 & 93.9 & 90.8 & 95.2 & 91.8 & 92.4 & 70.5\\
RAST \citep{hao-etal-2021-rast} & 77.8 & 72.5 & NA & 90.5 & 88.3 & 94.7 & 88.9 & 92.9 & 64.4\\
HCT \citep{DBLP:conf/aaai/JinSJ0G22} & 79.3 & 74.2 & NA & 92.7 & 90.2 & 94.4 & 89.3 & 93.5 & 65.3\\
RAU \citep{DBLP:conf/icassp/ZhangLWCX22} & NA & NA & NA & NA & 91.6 & NA & 90.6 & 93.9 & 68.4\\
QUEEN \citep{si-etal-2022-mining} & NA & NA & NA & NA & 92.1 & NA & 90.9 & 94.6 & 70.1\\
SGT \citep{chen-2023-incomplete} & 91.0 & \textbf{89.8} & 85.1 & 94.9 & 92.2 & 87.0 & 91.0 & 94.6 & 67.4\\
MIUR \citep{li-etal-2023-well} & NA & 82.2 & NA & NA & 91.2 & NA & 90.7 & 93.7 & 67.7\\
\midrule
EO-IUR (Ours) & \textbf{93.0} & 88.9 & \textbf{85.8} & \textbf{95.9} & \textbf{94.5} & \textbf{97.2} & \textbf{94.2} & \textbf{96.0} & \textbf{79.9}\\
\bottomrule
\end{tabular}
\caption{Result comparison on Chinese REWRITE.}
\label{tab:rewrite}
\end{table*}

\begin{table*}[]
\centering
\small
\setlength{\tabcolsep}{1.5mm}
\begin{tabular}{lcccccccccccccc}
\toprule
Model &  P$_1$  & R$_1$ & F$_1$ & P$_2$  & R$_2$ & F$_2$  & P$_3$  & R$_3$ & F$_3$ & B$_1$ & B$_2$ & R$_1$ &R$_2$ & EM \\
\midrule
BART$_{\operatorname{base}}$ \citep{lewis-etal-2020-bart} &  70.9 & 55.8	& 62.4	& 60.8	& 47.4	& 53.3	& 54.0	& 41.8	& 47.1	& 90.5	& 87.9	& 91.8	& 85.5 & 52.9\\
SARG \citep{DBLP:conf/aaai/HuangLZ021} & NA & NA & 62.4 & NA & NA & 52.5 & NA & NA & 46.3 & 92.2 & 89.6 & 92.1 & 86.0 & NA\\
RAU \citep{DBLP:conf/icassp/ZhangLWCX22} & 75.0 & 65.5 & 69.9 & 61.2 & 54.3 & 57.5 & 52.5 & 47.0 & 49.6 & 92.4 & 89.6 & 92.8 & 86.0 & NA\\
QUEEN \citep{si-etal-2022-mining} & NA & NA & NA & NA & NA & NA & NA & NA & NA & 92.4 & 89.8 & 92.5 & 86.3 & 53.5\\
MGIIF \citep{du-etal-2023-multi} & NA & NA & 70.8 & NA & NA & 58.5 & NA & NA & 50.5 & 93.1 & 90.4 & 93.2 & 86.6 & NA\\
MIUR \citep{li-etal-2023-well} & \textbf{76.4} & 63.7 & 69.5 & 62.7 & 52.7 & 57.3 & 54.3 & 45.9 & 49.7 & 93.0 & 90.1 & 92.6 & 85.7 & 51.0\\
\midrule
EO-IUR (Ours) & \textbf{76.4} & \textbf{69.1} & \textbf{72.5} & \textbf{67.4} & \textbf{60.5} & \textbf{63.7} & \textbf{61.2} & \textbf{54.7} & \textbf{57.8} & \textbf{93.5} & \textbf{91.2} & \textbf{93.4} & \textbf{87.9} & \textbf{59.1}\\
\bottomrule
\end{tabular}
\caption{Result comparison on Chinese RES200K.}
\label{tab:restoration}
\end{table*}

\begin{table}[h]
    \centering
     \setlength{\tabcolsep}{4pt}
    \resizebox{\linewidth}{!}{
    \begin{tabular}{lccc}
    \hline
         Method & EM & BLEU$_4$ & F$_1$  \\
         \hline
         \cellcolor[gray]{0.9} EO-IUR & \cellcolor[gray]{0.9} \textbf{80.8} & \cellcolor[gray]{0.9} \textbf{90.6} & \cellcolor[gray]{0.9} \textbf{87.6}\\
         \hline
          - w/o ED guidance & $\downarrow 3.8$ & $\downarrow 2.6$ & $\downarrow 2.6$\\
          - w/o Multi-task learning &  $\downarrow 4.4$ &  $\downarrow 2.1$ &  $\downarrow 2.8$\\
          - w/o Utterance augmentation & $\downarrow 1.2$ &  $\downarrow 0.4$ &  $\downarrow 0.6$\\
          - w/o Heterogeneous graph & $\downarrow 3.8$ & $\downarrow 2.0$ &  $\downarrow 1.5$\\
           \hline
          - One-hot label &  $\downarrow 3.6$ & $\downarrow 2.2$ &  $\downarrow 1.8$\\
          - Merge labels &  $\downarrow 2.5$ &  $\downarrow 1.2$ &  $\downarrow 1.4$\\
           \hline
    \end{tabular}
    }
    \caption{Ablation study on the TASK dataset.}
    \label{tab:ablation-rewrite}
\end{table}

\subsection{Experimental Results}
The experimental results are shown in Tables~\ref{tab:task}, ~\ref{tab:rewrite} and ~\ref{tab:restoration}. Our EO-IUR is significantly better in comparison with the best baselines in three datasets.
Due to the different purposes of the three datasets, we report the different metrics on the respective datasets following previous work \citep{chen-2023-incomplete,li-etal-2023-well}.
The better performance on BLEU$_n$ and ROUGE$_n$ demonstrates that our EO-IUR is capable of generating more accurate and less redundant utterances. In contrast to other text generation tasks, the content of incomplete and rewritten utterances in IUR is largely similar, making it challenging to discern the effects of utterance rewriting through the use of BLEU$_n$ and ROUGE$_n$. Consequently, the experimental results indicate that the performance improvement of these two metrics is not as significant as that observed in other metrics.

In comparison with BLEU$_n$ and ROUGE$_n$, our EO-IUR achieves more significant improvements in most F$_n$ and EM, with the EM metric increasing by 9.2, 9.4, and 5.6 on the TASK, REWRITE, and RES200K datasets, respectively.  The F$_n$ and EM metrics focus on the completion of coreferential and omitted information during the rewriting process. The substantial improvements achieved in these two types of metrics also demonstrate the effectiveness of our EO-IUR. It is noteworthy that the significant improvement in EM indicates that our EO-IUR can generate complete utterances.

QUEEN and MGIIF also focused on capturing key information, where the former employed manually designed rules to identify the pronouns to be replaced and the ellipsis in incomplete utterances, while the latter focuses on utterance-level importance information. In contrast to the aforementioned methods, we introduce token-level editing operations to measure the contribution of each token. The experimental results in Tables~\ref{tab:task}, ~\ref{tab:rewrite} and ~\ref{tab:restoration} show that our EO-IUR outperforms them significantly on all metrics, thereby substantiating the efficacy of our approach to selecting key tokens from the dialogue context at the token level.

\subsection{Analysis}
Due to the limited space available, we only reported the results on the TASK and the results are presented in Tables ~\ref{tab:ablation-rewrite}. The results on the other two datasets also demonstrated similar trends.

\noindent \textbf{Impact of Editing Operation-guided IUR}
When the editing operation guidance is removed (i.e., ``w/o ED guidance''), all of the model's metrics decrease significantly, demonstrating that this guidance can enhance utterance generation. Furthermore, when the entire editing operation labeling task is removed directly (i.e., ``w/o Multi-task learning''), including the editing operation guidance, there is a greater decrease in model performance, indicating that training on the sequence labeling task is also beneficial for utterance generation.

To comprehensively assess the efficacy of the editing operation-guided IUR in preventing the generation of redundant tokens during the rewriting process, we have collated the experimental results on TASK. In this analysis, we define a redundant token as one that does not exist in the reference utterances. Without the introduction of editing operation guidance to our model EO-IUR, 12.75\% of the generated rewritten utterances contain redundant tokens. This figure is reduced to 8.25\% after the incorporation of this guidance. This is due to the fact that the labels generated by editing operation labeling provide prior information to the generation model, allowing it to focus more on key tokens in dialogue context.

In utterance rewriting, the use of labels to guide the decoder is a crucial aspect of the process. These labels, which can be either soft or one-hot, serve as influencing factors that adjust attention scores in the cross-attention layers. In certain instances, one-hot labels can be directly utilized as influencing factors. 
As shown in Table ~\ref{tab:ablation-rewrite}, soft labels are demonstrably superior to one-hot labels (i.e., ``One-hot label'') as influencing factors. 
One potential issue with using one-hot labels during training is that it may lead to overfitting. Additionally, incorrectly predicting one-hot label during inference could affect the decoder's generation process. Another consideration is that, in addition to tokens corresponding to the ``RP'', ``NW'', and ``IN'' labels, some tokens corresponding to the ``NA'' label also require attention during generation. Therefore, solely focusing on the former may prevent the model from considering the complete semantic context of dialogues. Consequently, the utilization of soft labels can facilitate the acquisition of more comprehensive token information, effectively circumventing the aforementioned issues.

To assess the efficacy of our labeling approach,  we directly combine ``RP'', ``NW'' and ``IN'' into a single label ``ED'' (i.e., ``Merge labels''). As shown in Tables ~\ref{tab:ablation-rewrite}, our findings indicate that utilizing two labels results in a notable decline in performance. This phenomenon can be attributed to the fact that the four types of labels can provide more fine-grained information to distinguish different editing operations. ``RP'' and ``NW'' correspond to the replacement  operation in coreference resolution, while ``IN'' corresponds to the insertion operation in ellipsis resolution. These three labels can distinguish the cases of coreference resolution and ellipsis resolution in IUR.

\begin{figure}[t]
\begin{center}
 \includegraphics[width=1.05\linewidth]{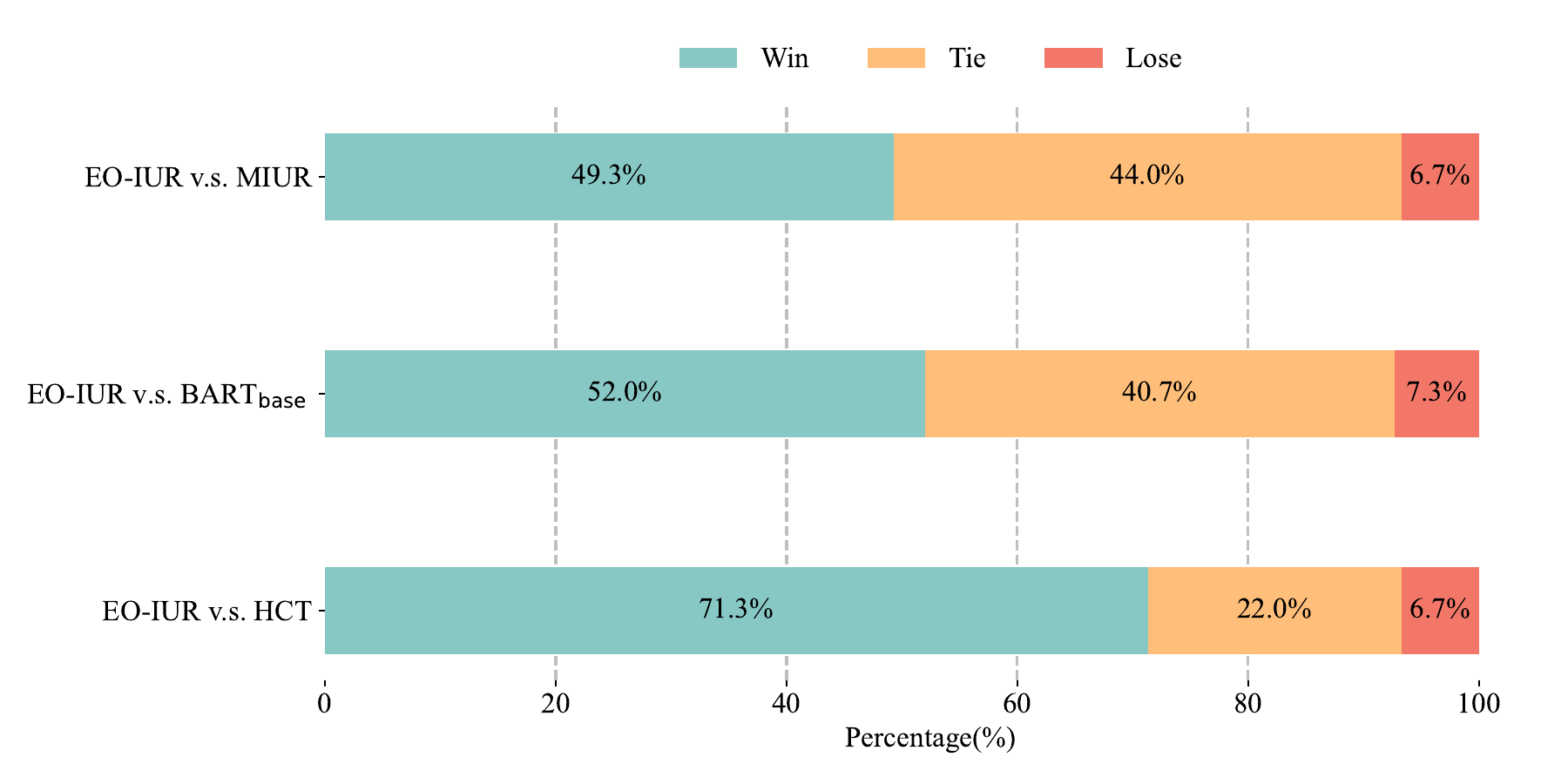}
 \caption{Human evaluation on REWRITE.}
 \label{fig:h_eval}
\end{center}
\vspace{-0.7cm}
\end{figure}

\noindent \textbf{Impact of Utterance Augmentation}
We conduct the experiments of removing utterance augmentation (i.e., ``w/o Utterance augmentation'') and all metrics drops in Table ~\ref{tab:ablation-rewrite}. 
The EM metric, which measures the proportion of samples that are completely correct after rewriting, demonstrates the most significant improvement after using data augmentation in comparison with the other metrics. However, existing methods may introduce minor errors or redundant tokens during the rewriting process, resulting in a lower EM score. As previously stated, the two data augmentation strategies employed herein result in the generation of a greater number of training samples, thereby enhancing the robustness of the model and facilitating the avoidance of errors to the greatest extent possible. 

\begin{table}[htbp]
\centering
\small
\resizebox{\linewidth}{!}{
\begin{tabular}{lcccccccc}
\toprule
Length & EM & F$_1$ & F$_2$ &	B$_1$ &	B$_2$ &	R$_1$ &	R$_2$ &	R$_L$   \\
\midrule
$[$1,5$]$	&	85.7	&	86.7	&	82.1	&	89.4	&	86.3	&	95.8	&	88.6	&	95.4\\
\midrule
$[$6,10$]$	&	72.2	&	86.3	&	81.9	&	91.5	&	89.9	&	94.9	&	91.6	&	94.1\\
\midrule
$[$11,15$]$	&	80.8	&	87.8	&	84.7	&	94.1	&	93.3	&	97.3	&	95.5	&	96.9\\
\midrule
$[$16,39$]$	&	87.0	&	90.6	&	89.4	&	97.2	&	96.7	&	98.3	&	97.3	&	98.0 \\

\bottomrule
\end{tabular}
}
\caption{Analysis of the correlation between model performance and incomplete utterance length.}
\label{tab:uttr_len}
\end{table}

\begin{table}[htbp]
\centering
\small
\resizebox{\linewidth}{!}{
\begin{tabular}{ccccccccc}
\toprule
Number & EM & F$_1$ & F$_2$ &	B$_1$ &	B$_2$ &	R$_1$ &	R$_2$ &	R$_L$   \\
\midrule
0	&	97.9	&	98.7	&	98.2	&	99.6	&	99.3	&	99.8	&	96.8	&	99.7\\
\midrule
1	&	62.6	&	84.2	&	80.1	&	89.4	&	87.7	&	93.0	&	88.7	&	92.1\\
\midrule
2	&	62.6	&	84.3	&	80.5	&	90.1	&	88.2	&	92.9	&	88.4	&	91.7\\
\midrule
3	&	25.0	&	67.5	&	57.8	&	68.8	&	63.7	&	78.9	&	66.5	&	77.5 \\
\midrule
4	&	20.0	&	74.0	&	61.5	&	68.2	&	64.4	&	81.6	&	70.7	&	78.5 \\
\midrule
5	&	0	&	80.0	&	70.6	&	71.7	&	67.0	&	85.7	&	73.7	&	82.4 \\
\bottomrule
\end{tabular}
}
\caption{Analysis of the correlation between model performance and the number of editing operations.}
\label{tab:ed_num}
\end{table}

\noindent \textbf{Impact of the Length of Incomplete Utterances and the Number of Editing Operations}
Taking the TASK dataset as an example, we evaluate the performance of the model when the incomplete utterances are in different lengths, as shown in Table~\ref{tab:uttr_len}. It can be observed that the model demonstrates superior performance when the incomplete utterances are of shorter or longer duration. However, it exhibits diminished performance when the length of the incomplete utterances falls within the interval [6, 10] and the interval [11, 15]. We analyzed the data and found that most short utterances (such as "Thank you. Goodbye.") lacked meaningful content and most long utterances were generally complete, which did not need to be rewritten. Incomplete utterances are primarily observed in utterances of intermediate length, specifically within the intervals [6, 10] and [11, 15], which present a greater challenge.

We also analyzed the correlation between the number of editing operations in the utterances and the model performance, as shown in Table~\ref{tab:ed_num}. It can be observed that as the number of editing operations increases, the model's performance decreases. For example, when the number of editing operations is 3, 4, and 5, the EM scores are 25.0, 20.0, and 0, respectively. How to address incomplete utterance rewriting in complex scenarios is also a research direction worthy of exploration in the future.

We also analyze the impact of token-level heterogeneous graph and provide the case study in the Appendix~\ref{appendix:graph} and~\ref{appendix:case}.

\subsection{Human Evaluation}

\begin{table}[t]
\centering
\resizebox{\linewidth}{!}{
\begin{tabular}{lccccccccccccc}
\toprule

Model & F$_1$ & F$_2$  & B$_1$ & B$_2$ & R$_1$ &R$_2$ & R$_L$ & EM   \\
\midrule

GPT-4 & 79.1	& 66.7	& 85.3	& 80.4	& 87.7	& 77.6	& 82.1	& 35.7\\
\midrule
EO-IUR & \textbf{93.0} & \textbf{88.9} & \textbf{95.9} & \textbf{94.5} & \textbf{97.2} & \textbf{94.2} & \textbf{96.0} & \textbf{79.9}\\

\bottomrule
\end{tabular}
}
\caption{Performance comparison to GPT-4 on REWRITE.}
\label{tab:gpt-comp}
\end{table}
As the rewritten results in IUR are often not unique and it is difficult to fully reflect the performance based on automatic evaluation metrics alone, a manual pair-wise evaluation was conducted to compare our proposed method with the strong baselines HCT, BART$_{\operatorname{base}}$ and MIUR. A random sample of 50 pieces of data from the REWRITE dataset was distributed to three raters (graduate students in NLP), who were asked to select the result generated by the two methods that was of superior quality. As shown in Figure~\ref{fig:h_eval}, our method outperforms the other three models in human evaluation significantly, further demonstrating its effectiveness. Furthermore, it is noteworthy that BART$_{\operatorname{base}}$ and our method exhibit comparable performance in many instances, and the failure rate of our EO-IUR is highest among the comparison with three models. Our observations indicated that while BART$_{\operatorname{base}}$ generated numerous utterances that did not meet rewrite requirements, their fluency led to superior manual evaluation results. In contrast, HCT and MIUR generated many utterances with grammatical errors due to the use of sequence labeling methods, resulting in poor manual evaluation performance.

\subsection{Comparison with ChatGPT}

The majority of evaluations of LLMs currently focus on single-turn dialogues, with minimal attention paid to multi-turn dialogues. However, the IUR tasks necessitate a comprehensive grasp of the global and structural information present in multi-turn dialogues. To investigate the performance of LLMs in IUR, we have conducted a comparative analysis between ChatGPT and our EO-IUR. Our approach employs the in-context learning approach, provides prompts and five examples, and utilises GPT-4 (the version used is \texttt{gpt-4-1106-preview}) to generate rewritten utterances. We have provided the prompt in Appendix~\ref{prompt}.

The experimental results are shown in Table~\ref{tab:gpt-comp}. Despite GPT-4 having a significantly greater number of parameters than our EO-IUR, our model still achieves superior performance under these conditions, particularly in terms of EM, which is almost twice as much as GPT-4. According to our observation, the results generated by GPT-4 suffer from "under-rewriting" and hallucination, that is, the rewriting is incomplete and some utterances are generated that is contradictory to the conversation context, resulting in low performance. We provide a comparison of the rewritten utterances generated by our model and GPT-4 in Appendix~\ref{comp_samples}.

\subsection{Error Analysis}
To conduct a more in-depth study of our proposed model, we have statistically analyzed the experimental results and conducted an error analysis. Due to our use of the sum of the probabilities that each input token belongs to the labels “RP”, “NW” and “IN” as an influence factor to guide the generation of rewritten utterances, we speculate that the performance of editing operation labeling will greatly affect the generation of rewritten utterances. Consequently, the performance of editing operation labeling was evaluated using the REWRITE dataset as an illustrative example, with the EM metric employed as the evaluation metric. The results revealed that the EM metric for editing operation labeling was 76.35, which was lower than that for rewritten utterance generation (79.9). This indicates that approximately a quarter of the generated labels were incorrect, which has a significant impact on subsequent utterance generation. Nevertheless, this also indicates that correct rewritten utterances can still be generated even when some labels in certain utterances are predicted incorrectly. It also demonstrates that the use of soft labels can effectively alleviate the problem of model overconfidence and improve the robustness of our model. Conversely, other errors originate from utterance generation itself. Despite the guidance of the correct labels, the utterance generation model still generates incomplete utterances, utterances with grammatical errors, utterances with redundant tokens. 

\section{Conclusion}
In this paper, we introduce the editing operation labels generated by the editing operation labeling to guide the generation model to focus on critical tokens. Furthermore, we introduce a token-level heterogeneous graph to learn the syntactic structure and the relationships among coreference-related tokens. We also propose a two-dimensional utterance augmentation strategy to further boost the model. The experimental results on three datasets show that our EO-IUR outperforms previous SOTA baselines significantly. In the future, we intend to employ large language models (LLMs) to facilitate utterance generation and to reduce the number of redundant tokens in rewritten utterances.

\section*{Limitations}
Although our proposed method effectively integrates the generation task with the classification task to leverage the strengths of both, there are still some drawbacks. Firstly, incorporating information from heterogeneous graphs of dialogues via graph neural networks incurs additional computational resources. On the other hand, in the process of using classification results to guide the generation of rewritten utterances, there are errors in the classification results, which may have a negative impact on generation. Therefore, it is worthwhile to explore how to efficiently integrate graph information and better combine classification and generation in future research.

\section*{Acknowledgements}
The authors would like to thank the three anonymous reviewers for their comments on this paper. This research was supported by the National Natural Science Foundation of China (Nos. 62276177 and 62376181), and Project Funded by the Priority Academic Program Development of Jiangsu Higher Education Institutions.

\bibliography{custom}

\appendix

\section{Examples of Data Augmentation}
\label{appendix:augexample}
We provide an example of data augmentation in Table ~\ref{tab:LLM_prompt}.
\begin{table}[!ht]
  \caption{The example of the two-dimensional data augmentation method we proposed.
  }
\small
  \centering
\resizebox{\linewidth}{!}{
    \begin{tabularx}{\linewidth}{X}
    \toprule
    \rowcolor[gray]{0.95}\multicolumn{1}{c}{\textbf{I: Initial dialogue}} \\
    \makecell[l]{
    \color{gray}{/* \textit{Dialogue history} */}\\
    \textbf{Speaker1}: Xiaowei, is there anything fun in Qingdao? \\
    \textbf{Speaker2}: I think there is a place in Qingdao that you \\must visit, the Badaguan Scenic Spot.\\
    \color{gray}{/* \textit{Source utterance} */} \\
    \textbf{Speaker1}: Please provide a specific introduction.\\
    \color{gray}{/* \textit{Target utterance} */} \\
    \textbf{Speaker1}: Please provide a detailed introduction to the \\ Badaguan Scenic Spot.}\\
    \midrule
    \rowcolor[gray]{0.95}\multicolumn{1}{c}{\textbf{II: Dialogue beyond data augmentation}} \\
    \makecell[l]{\color{gray}{/* \textit{Dialogue history} */}\\
    \textbf{Speaker1}: Do you know where is the best place to go in \\ Qingdao? \\
    \textbf{Speaker2}: I recommend that you go to the Badaguan \\Scenic Spot to experience it.\\
    \color{gray}{/* \textit{Source utterance} */} \\
    \textbf{Speaker1}: Please provide a specific introduction (to) it.\\
    \color{gray}{/* \textit{Target utterance} */} \\
    \textbf{Speaker1}: Please provide a detailed introduction to the \\Badaguan Scenic Spot.}\\
    \bottomrule
    \end{tabularx}
    }
  \label{tab:LLM_prompt}
\end{table}

\section{Prompt used in LLM-based Historical Utterance Augmentation}
\label{appendix:augprompt}
The following prompt is used to instruct LLM to rewrite historical utterances,
\begin{tcolorbox}[colback=blue!5!white,colframe=Emerald!80!black,title=Prompt for historical utterance augmentation]
  Given a dialogue with utterances from different speakers separated by semicolons, keep the last utterance unchanged, rewrite the historical utterances, and keep the semantics of the dialogue unchanged. Here are some examples. \\
  Examples: \{\texttt{Examples}\} \\
  Input: \{\texttt{Input}\}
\end{tcolorbox}
\noindent where ``\texttt{Examples}'' are the five highest-quality examples we have selected, and the ``\texttt{Input}'' is the historical utterances that needs to be augmented, accompanied by its corresponding incomplete utterance. To illustrate, the utterance ``Is there anything fun in Qingdao?'' can be rewritten by LLM as ``Do you know where is the best place to go in Qingdao?'' with explicit semantics in a dialogue about tourism.

\section{Dataset statistics}
\label{appendix:dataset}
The specific information of the three datasets is shown in Table ~\ref{tb:data_stat}, where ``Avg. Hist'', ``Avg. Curr'' and ``Avg. Rewr'' refer to the average numbers of tokens in the historical utterances, current utterances, and rewritten utterances, respectively. It can be observed that the rate of replacement operations in REWRITE is considerably higher (35.8\%) than in RES200K (0.15\%) and TASK (12.4\%). Additionally, the average number of tokens in historical utterances of TASK is significantly larger than those in the other two datasets.
\begin{table}[h]
\centering
\small
\resizebox{\linewidth}{!}{
\begin{tabular}{cccc}
\toprule
Category & REWRITE & RES200K & TASK \\
\midrule
Language & Chinese & Chinese & English \\
Train & 18K & 194K & 2.2K \\
Dev & 2K & 5K & 0.5K \\
Test & 2K & 5K & 0.5K \\
\#Avg. Hist & 17.7 & 25.5 & 52.6 \\
\#Avg. Curr & 6.5 & 8.6 & 9.4 \\
\#Avg. Rewr & 10.5 & 12.4 & 11.3 \\
\#Insertion & 14070 & 136339 & 1572 \\
\#Replacement & 7853 & 203 & 223 \\
\bottomrule
\end{tabular}
}
\caption{Statistics of different datasets.}
\label{tb:data_stat}
\end{table}

\section{Implementation Details}
\label{appendix:implementation}
For fair comparison, we use BART$_{\operatorname{base}}$ as the backbone model and employ a 6-layer convolutional neural network for feature extraction. The number of epochs for warm-up is set to 3. In multi-task learning, we set $\alpha_1=1$, $\alpha_2=1$. For the temperature coefficient $\tau_d$, we set it to 1. We use \texttt{gpt-3.5-turbo}\footnote[1]{\url{https://chat.openai.com/}} for historical utterance augmentation via the OpenAI API.

\section{Impact of Token-level Heterogeneous Graph}
\label{appendix:graph}
When the token-level heterogeneous graph (i.e., ``w/o heterogeneous graph'') is removed, the BART encoder is employed solely to represent dialogues. As illustrated in Table ~\ref{tab:ablation-rewrite}, the metrics demonstrate a decline. This outcome indicates that the structural information within dialogues can link the coreference-related tokens in a dialogue and reveal the ellipsis tokens in incomplete utterances, subsequently enhancing the performance of IUR. 
Additionally, experiments were conducted on four types of edges, and the results demonstrated that removing any type of edges would result in a performance decrease. Specifically, removing intra-utterance edges, inter-utterance edges, speaker-utterance edges, and pseudo-coreference edges reduced the EM metric by 1.4, 0.5, 0.2, and 0.9, respectively. These findings further substantiate the effectiveness of these four types of edges. 

We observe that the removal of intra-utterance edges and pseudo-coreference edges results in the largest decrease relative to the removal of other types of edges, which highlights the importance of these two types of edges. This is mainly due to the fact that the intra-utterance edges incorporate syntactic information well, allowing the model to better grasp what syntactic components are missing in the incomplete utterance, which often corresponds to omitted content. Furthermore, pseudo-coreference edges can effectively fuse coreference information in the dialogue. Additionally, the presence of numerous personal pronouns in the dialogue allows for the use of such edges, which can enhance the model's awareness of the coreference relations in the dialogue.

\section{Case Study}
\label{appendix:case}
\begin{table*}[!ht]
    \centering
     \resizebox{\linewidth}{!}{
    \begin{tabular}{ll}
    \toprule
    \textbf{Dialogue context} & \begin{tabular}[c]{@{}l@{}} \emph{$Speaker_1$: How about indian food \textcolor[RGB]{180,122,132}{in the north part of town} instead?}\\ \emph{$Speaker_2$: There are two \textcolor[RGB]{248,122,23}{indian} restaurants in the north part of town . what price range are you looking for?}\\ \emph{$Speaker_1$: What are the price ranges of the two \textcolor[RGB]{248,122,23}{available} restaurants? (\color{blue}{The incomplete utterance.})} \end{tabular} \\
    \midrule
         \textbf{Reference utterance} & \emph{What are the price ranges of the two \textcolor[RGB]{248,122,23}{indian} restaurants \textcolor[RGB]{180,122,132}{in the north part of town?}} \\ 
            \midrule
         \textbf{MIUR} & \emph{What are the price ranges of the two available restaurants north of?} \xmark \\ 
         \midrule
         \textbf{BART$_{\operatorname{base}}$} & \emph{What are the price ranges of the two indian restaurants?} \xmark \\ 
         \midrule
         \textbf{EO-IUR (Ours)} & \emph{What are the price ranges of the two indian restaurants in the north part of town?} \cmark \\
    \bottomrule
    \end{tabular}
    }
    \caption{A case study of our proposed model, MIUR, and BART$_{\operatorname{base}}$ on the TASK dataset.}
    \label{tab:case-study}
\end{table*}

\begin{table*}[!ht]
    \centering
    \small 
    \captionsetup{justification=raggedright}
    \setlength{\tabcolsep}{10pt}
    \begin{tabular}{@{}p{8.5cm}m{3cm}m{3cm}@{}} 
    \hline
    \textbf{Dialogue Context} & \textbf{GPT-4} & \textbf{EO-IUR (Ours)} \\ 
    \hline
    \begin{tabular}[l]{p{8.5cm}} A: How about an expensive restaurant that serves korean food?\\ B: I apologize, there are no restaurants in the expensive price range that serve welsh food. Would you like a different type of food?\\ A: How about korean food? (\color{blue}{The incomplete utterance.}) \end{tabular} & How about an expensive restaurant that serves korean food? & How about korean food in the expensive price range? \\
    \hline
    \begin{tabular}[l]{p{8.5cm}} A: Tell me the address of the moderately priced indian restaurant.\\ B: The nirala is located at 7 milton street chesterton.\\ A: What is the phone number? (\color{blue}{The incomplete utterance.}) \end{tabular} & What is the phone number of the moderately priced indian restaurant? & What is the phone number of the nirala? \\
    \hline
    \begin{tabular}[l]{p{8.5cm}} A: I want to find a cheap restaurant and serves australasian food.\\ B: There are no restaurants serving australasian food. Would you like something else?\\ A: Portuguese food. (\color{blue}{The incomplete utterance.}) \end{tabular} & Portuguese food. & I would like cheap portuguese food.\\
    \hline
    \end{tabular}
    \caption{Comparison of examples generated by GPT-4 and our EO-IUR.}
    \label{comp_examples}
\end{table*}

We conducted a case study to further explore the effectiveness of our proposed method, as shown in Table ~\ref{tab:case-study}. In the incomplete utterance, ``available restaurants'' refers to ``indian restaurants'', and ``in the north part of town'' is omitted. The phrase ``in the north part of town'' is a prepositional phrase modifying ``the two indian restaurants''. In order for the model to correctly rewrite this utterance, it must first learn the coreference relation between ``available restaurants'' and ``indian restaurants'', as well as the syntactic structure that ``in the north part of town'' modifies ``the two indian restaurants''.

We compared the results generated by MIUR, BART$_{\operatorname{base}}$ and our proposed model. Only our method generated a correct rewritten utterance, while MIUR generated an utterance with grammatical error due to the inherent flaws of sequence labeling methods that make it difficult to guarantee grammatical correctness. Furthermore, MIUR also did not correctly identify the coreference relation between ``available restaurants'' and ``indian restaurants''. The utterance generated by BART$_{\operatorname{base}}$ is incomplete and lacks the modifier ``in the north part of town''. 
Our method benefits from the editing operation guidance and heterogeneous graph representation of dialogue, enabling it to generate a correctly rewritten utterance. This is due to the ability of our EO-IUR to not only learn the coreference relation but also to capture the syntactic structure. It is noteworthy that our EO-IUR correctly identifies the tags ``RP'' and ``NW'' for ``available'' and ``Indian'', respectively. This enables the model to prioritize these tokens during the generation of rewritten utterance.

\section{Prompts Used in GPT-4 Evaluation}
\label{prompt}
The prompt used in the GPT-4 evaluation is as follows.
\begin{tcolorbox}[colback=Emerald!9,colframe=cyan!40!black,title=Prompt used in GPT-4 assessment]
  The goal of dialogue rewriting is to resolve coreference and ellipsis, that is, to complete the coreferential and omitted information in the dialogue without changing its original semantics. Please rewrite the final utterance in the following dialogue. \\
  Examples: \{\texttt{Examples}\} \\
  Input: \{\texttt{Input}\}
\end{tcolorbox}

\section{Sample Comparative Analysis between GPT-4 and EO-IUR}
\label{comp_samples}

Three examples are given in Table~\ref{comp_examples} where GPT-4 rewrites incorrectly but our method rewrites correctly. For the first example, speaker A asks for Korean food, while GPT-4 rewrites it to ask for an expensive restaurant, which we speculate is due to the fact that GPT-4 overly relies on speaker A's first sentence (i.e., "I am looking for an expensive restaurant that serves welsh food."), resulting in the output of an incorrectly rewritten utterance. For the second example, speaker A asked for the phone number of nirala, while the utterance output by GPT-4 asked for the phone number of "moderately priced Indian restaurant". For the third example, GPT-4 does not complete the omitted content of the utterance.

\end{document}